\documentclass{article}
\usepackage{spconf,amsmath,epsfig,amssymb,xcolor,bbm,theorem,enumerate,graphicx}
\usepackage[linesnumbered,vlined,boxed,commentsnumbered]{algorithm2e}
\ninept

\def\nset{{\mathbb{N}}}
\def\rset{\mathbb R}

\def\Zset{\mathsf{Z}}
\def\Zsigma{\mathcal{Z}}

\def\rmd{\mathrm{d}}

\def\1{\mathbbm{1}}

\def\PP{\mathbb{P}} 
\def\PE{\mathbb{E}} 

\newcommand{\pscal}[2]{\left\langle#1,#2\right\rangle}
\def\Id{\mathrm{I}}
\newcommand{\eqdef}{\ensuremath{\stackrel{\mathrm{def}}{=}}}

\newcommand{\kouter}{k_\mathrm{out}}
\newcommand{\kin}{k_\mathrm{in}}

\newcommand\curr{\mathrm{curr}}
\newcommand\init{\mathrm{init}}
\newcommand{\R}{\mathsf{R}}

\newcommand{\bars}{\bar{\mathsf{s}}}
\newcommand{\s}{\mathsf{s}}
\newcommand{\hats}{\hat{\mathsf{s}}}
\newcommand{\mf}{\mathsf{h}}

\newcommand{\hatS}{\widehat{S}}
\newcommand{\Smem}{\mathsf{S}}

\newcommand{\Sset}{\mathcal{S}}

\newcommand{\Prox}{\mathrm{Prox}}

\newcommand{\param}{\theta}
\newcommand{\Param}{\Theta}
\newcommand{\map}{\mathsf{T}}
\newcommand{\lyap}{\operatorname{W}}
\newcommand{\batch}{\mathcal{B}}
\newcommand{\lbatch}{\mathsf{b}}
\newcommand{\pas}{\gamma}

\def\indic{\chi}

\def\eqsp{\;}

\title{The Perturbed Prox-Preconditioned SPIDER algorithm for EM-based
large scale learning} \name{G. Fort$^1$, E. Moulines$^2$ \thanks{This
      work is partially supported by the {\em Fondation Simone et Cino
        Del Duca} through the project OpSiMorE and by the ANR-19-CHIA-0002-01. Part of this work was conducted under the auspices of the Lagrange Center in Mathematics and Computer Sciences}} \address{$^1$ IMT,
    Universit\'e de Toulouse \& CNRS, F-31062 Toulouse, France.
    \\ $^2$ CMAP, Ecole Polytechnique, Route de Saclay, 91128
    Palaiseau Cedex, France.  }

\begin{document}
\ninept
\maketitle
\begin{abstract}
      Incremental Expectation Maximization (EM) algorithms were
      introduced to design EM for the large scale learning framework
      by avoiding the full data set to be processed at each
      iteration. Nevertheless, these algorithms all assume that the
      conditional expectations of the sufficient statistics are
      explicit. In this paper, we propose a novel algorithm named {\tt
        Perturbed Prox-Preconditioned SPIDER} (3P-SPIDER), which
      builds on the {\tt Stochastic Path Integral Differential
        EstimatoR EM} (SPIDER-EM) algorithm. The {\tt 3P-SPIDER}
      algorithm addresses many intractabilities of the E-step of EM;
      it also deals with non-smooth regularization and convex
      constraint set. Numerical experiments show that {\tt 3P-SPIDER}
      outperforms other incremental EM methods and discuss the role of
      some design parameters.
\end{abstract}
\section{Introduction}
\label{sec:intro}
EM~\cite{Dempster:em:1977,Wu:1983} is a very popular computational
tool, designed to solve non convex minimization problems on $\rset^d$
when the objective function is not explicit but defined by an integral
$F(\theta) = -\log \int_\Zset G(z;\theta) \rmd \mu(z)$.  EM is a
Majorize-Minimization algorithm which, based on the current value of
the parameter $\theta_\curr$, defines a majorizing function $\theta
\mapsto Q(\theta;\theta_\curr)$ through a Kullback-Leibler argument;
then, the new point is chosen as the/a minimum of
$Q(\cdot;\param_\curr)$. The computation of $Q$ is straightforward
when there exist (known and explicit) functions $\R,\phi,\s$ such that
$Q(\cdot; \theta_\curr) = \R(\cdot) -
\pscal{\bars(\theta_\curr)}{\phi(\cdot)}$ and $\bars(\tau) \propto
\int_\Zset \s(z) G(z;\tau) \rmd \mu(z)$ is the expectation of the
function $\s$ with respect to (w.r.t.) the probability measure
$G(\cdot; \tau) \exp(-F(\tau))\rmd \mu$. In these cases, the vector
$\bars(\theta_\curr)$ defines the function $Q$.

It may happen that the vector $\bars(\theta_\curr)$ is not explicit
(see e.g. \cite[section 6]{maclachlan:2008}); a natural idea is to
substitute $\bars$ for an approximation, possibly random. A first
level of intractability occurs when the integral $\bars(\param_\curr)$
is not explicit.  Many stochastic EM versions were proposed and
studied to overcome this intractability: among them, let us cite Monte
Carlo EM \cite{Wei:tanner:mcem,Fort:moulines:2003} where $\bars$ is
approximated by a Monte Carlo integration; and SA EM
\cite{celeuxd85,Delyon:lavielle:moulines:1999} where $\bars$ is
approximated by a Stochastic Approximation (SA)
scheme~\cite{benveniste:etal:1990}.  With the Big Data era, a second
level of intractability occurred: EM applied to statistical learning
evolved into online versions and large scale versions in order to
minimize a loss function associated to a set of observations (also
called {\em examples}). In large scale versions, the number of
training data $n$ is too large to be processed at each iteration of
EM: for example, when the majorizing function $Q$ of EM is of the form
$Q(\param; \param_\curr) = \R(\param) -
\pscal{\bars(\param_\curr)}{\phi(\param)}$, the vector
$\bars(\param_\curr)$ often has the form $n^{-1} \sum_{i=1}^n
\bars_i(\param_\curr)$ and the sum over $n$ terms can not be allowed
at each iteration of EM.  To overcome this intractability in this
so-called {\em finite-sum} setting, incremental EM-based algorithms
were proposed: let us cite {\tt incremental EM}
\cite{Neal:hinton:1998}, {\tt Online-EM} \cite{cappe:moulines:2009},
     {\tt sEM-VR} \cite{chen:etal:2018}, {\tt FIEM}
     \cite{karimi:etal:2019} (see also \cite{fort:moulines:gach:2020}
     for {\tt opt-FIEM}) and {\tt SPIDER-EM}
     \cite{fort:moulines:wai:2020,fort:moulines:wai:2021}. The three
     algorithms {\tt sEM-vr}, {\tt FIEM} and {\tt SPIDER-EM} can be
     seen as a {\tt Online-EM} algorithm combined with a variance
     reduction technique through the construction of a control
     variate; they all improve on {\tt Online-EM} (see
     e.g. \cite{fort:moulines:wai:2021}). However, these EM-based
     algorithms designed for the {\em finite-sum} framework all
     consider that the functions $\param \mapsto \bars_i(\param)$ can
     be explicitly evaluated for any $\param$ and $i=1, \cdots, n$,
     while being defined as an expectation.

This paper introduces a novel EM-based procedure, named {\tt Perturbed
  Prox-Preconditioned SPIDER} which tackles the two difficulties: ;
\text{(i)} the {\em finite-sum} setting; \text{(ii)} the
intractability of the quantities $\bars_i(\param_\curr)$.  It is
proved in \cite{fort:moulines:wai:2020} that the complexity bounds of
{\tt SPIDER-EM}, expressed as the number of optimization steps and as
the number of evaluations of the quantities $\bars_i(\param_\curr)$
required to reach an $\epsilon$-approximate stationary point of $F$,
improves over the state-of-the art. Therefore, our algorithm builds on
{\tt SPIDER-EM}.  It is also designed to address a composite problem
with a non-smooth term.  {\tt 3P-SPIDER} is introduced in
Section~\ref{sec:notations:assumptions}, with an emphasis on the case
the quantities $\bars_i(\param_\curr)$ are approximated by a Monte
Carlo sum. In Section~\ref{sec:logistic}, the algorithm is applied to
the logistic regression problem; insights on the choice of some design
parameters are also given. It is shown that this perturbed version of
{\tt SPIDER-EM} improves on the perturbed version of {\tt Online-EM}
thus illustrating that the variance reduction technique is still
perceptible. This benefit is all the more visible that the error when
approximating the $\bars_i(\param_\curr)$'s is small. Finally, since
{\tt 3P-SPIDER} combines two approximations to address the
intractability of the $\bars_i(\param_\curr)$'s and the finite-sum
setting, it is advocated to regularly refresh the control-variate
approximation with a full screening of the data set.

The complexity analysis of this algorithm is provided in
\cite{fort:moulines:SSP21:theory}: under conditions on the
approximations of the $\bars_i(\param_\curr)'s$, which are satisfied
for example for a Monte Carlo approximation, it is shown that {\tt
  3P-SPIDER} has the same complexity bounds as {\tt SPIDER-EM}. In
that sense, it remains optimal among the (perturbed) incremental EM
algorithms.

{\bf Notations} $\rset_+^\star$ and $\nset^\star$ denote respectively
(resp.) the positive real line and the set of the positive
integers. For $n \in \nset^\star$, set $[n]^\star \eqdef \{1, \cdots,
n\}$ and $[n] \eqdef \{0, \cdots, n\}$.  For $x \in \rset$, $\lceil x
\rceil$ is the nearest integer greater than or equal to $x$. Vectors
are column-vectors; for $a,b$ in $\rset^\ell$, $\pscal{a}{b}$ denotes
the Euclidean scalar product, and $\|a\|$ the associated norm. For a
matrix $A$, $A^T$ and $A^{-1}$ are resp. its transpose and its
inverse. $\Id_d$ is the $d \times d$ identity matrix.  The random
variables are defined on a probability space $(\Omega, \mathcal{A},
\PP)$; $\PE$ denotes the associated expectation. For random variables
$U,V$, $\PE[U \vert V]$ is the conditional expectation of $U$ given
$V$. For a smooth function $f$, $\nabla_x f$ (or simply $\nabla f$
when clear enough) is the gradient of $f$ with respect to the variable
$x$; $\nabla^2 f$ is its hessian. For a proper lower semi-continuous
convex function $g$ and $x$ in its (assumed) non-empty domain,
$\partial g(x)$ is the subdifferential of $g$ at $x$.

\vspace{-0.2cm}

\section{The Perturbed Prox-Preconditioned SPIDER algorithm}
\label{sec:notations:assumptions}
\subsection{The optimization problem}
\label{sec:optimization}
We address the minimization of an objective function $F: \Theta\to
\rset$:
\begin{equation}
  \label{eq:problem}
 \param \mapsto \frac{-1}{n} \sum_{i=1}^n \log \int_{\Zset} h_i(z)
 \exp( \pscal{\s_i(z)}{\phi(\param)}) \rmd \mu(z) + \R(\param)
 \vspace{-0.3cm}
\end{equation}
where $\Param$ is an open subset of $\rset^d$, $(\Zset, \Zsigma)$ is a
measurable space, $\Zsigma$ denoting a $\sigma$-algebra over $\Zset$;
the functions $\phi:\Param \to \rset^q$, $\R: \Param \to \rset$ and
for all $i \in [n]^\star$, $\s_i:\Zset \to \rset^q$ and $ h_i: \Zset
\to \rset_+^\star$ are measurable; and $\mu$ is a dominating measure
on $(\Zset, \Zsigma)$.  The minimization of the negative
log-likelihood in latent variable models provides examples of such a
problem. As a first example, consider the maximum likelihood estimate
of a mixture of densities from the curved exponential family (see
e.g. \cite[supp. material]{fort:moulines:wai:2021} for the Gaussian
mixture model).  As a second example, consider the following logistic
regression model: given $\rset^d$-valued covariate vectors $\{X_i, i
\in [n]^\star \}$, for any $\param \in \Theta \eqdef \rset^d$, the
binary observations $\{Y_i, i \in [n]^\star \}$ are independent with
distribution
\begin{multline*}
  \vspace{-0.3cm}
  p_\param(y_i) \propto \int_{\rset^d} (1+\exp(-y_i
\pscal{X_i}{z_i}))^{-1} \\ \times \exp\left(-(2 \sigma^2)^{-1} \| z_i
- \theta \|^2 \right) \rmd z_i \eqsp,
\end{multline*}
for any $i \in [n]^\star$, $y_i \in \{-1,1\}$. In words, each
individual $\# i$ in the training set has an individual predictor
$Z_i$. Given $Z_i$, the success probability $\PP(Y_i = 1 \mid Z_i)$ is
$(1+\exp(-\pscal{X_i}{Z_i}))^{-1}$. The individual predictors $Z_1,
\cdots, Z_n$ are assumed to have a Gaussian distribution with
expectation $\theta$, assumed to be unknown, and (known) diagonal
covariance matrix $\sigma^2 \Id_d$. The ridge-regularized negative
log-likelihood, given by $-n^{-1} \sum_{i=1}^n \log p_\theta(Y_i) +
\tau \|\theta\|^2$ may be written as \eqref{eq:problem} with $\Zset
\eqdef \rset$, $\phi(\theta) \eqdef \theta$, $\rmd \mu(z) \eqdef \exp(
- z^2/(2\sigma^2)) \rmd z$, \vspace{-0.1cm} {\small \begin{align*} h_i(z) & \eqdef
    \left( 1 + \exp\left(-Y_i \|X_i\| z \right) \right)^{-1} \eqsp,
    \quad \s_i(z) \eqdef z \, \frac{X_i}{\sigma^2 \|X_i\|} \eqsp,
    \\ \R(\param) & \eqdef \frac{1}{2} \param^T \left(
    \frac{1}{\sigma^2 n} \sum_{i=1}^n \frac{X_iX_i^T}{\|X_i\|^2} + 2
    \tau \Id_d \right) \param \eqsp.
       \end{align*}}
\vspace{-0.6cm}
\subsection{EM in the expectation space}
For solving this optimization problem, EM defines a sequence
$\{\param_k, k \geq 0 \}$ taking values in $\Param$, by repeating
\textit{(i)} E-step: compute
\vspace{-0.2cm}
\[
Q(\param; \param_k) \eqdef - \frac{1}{n} \sum_{i=1}^n \int_\Zset
\pscal{\s(z)}{\phi(\param)} \, p_i(z; \param_k) \rmd \mu(z)
+\R(\param)
\]
where for any $z \in \Zset$, $\param \in \Param$, $i \in [n]^\star$,
\begin{equation}
\label{eq:aposteriori}
p_i(z;\param) \propto h_i(z) \exp(
\pscal{\s_i(z)}{\phi(\param)})
\end{equation}
is a probability density; \textit{(ii)} M-step: compute the minimum
\[
\param_{k+1} \eqdef \mathrm{argmin}_{\param \in \Param} Q(\param;
\param_k), \quad Q(\param; \param_k) = \R(\param) -
\pscal{\bars(\param_k)}{\phi(\param)},
\]
with
\[
\bars(\param)  \eqdef \frac{1}{n} \sum_{i=1}^n \bars_i(\param),
\qquad \bars_i(\param) \eqdef \int_\Zset \s_i(z) \, p_i(z;\param)
\rmd \mu(z) \eqsp.
\]
Hereafter, we assume that for any $\param_\curr \in \Param$, $\param
\mapsto Q(\param;\param_\curr)$ possesses an unique minimum and we
define for any $s$ in a closed convex set $\Sset \supseteq
\bars(\Param)$,
\[
\map(s) \eqdef \mathrm{argmin}_{\param \in \Param} \left( \R(\param) -
\pscal{s}{\phi(\param)} \right) \eqsp.
\]
With these notations, it holds: $\param_{k+1} = \map \circ
\bars(\param_k)$.

In the logistic regression example, $p_i(z; \param) \rmd \mu(z)$ is
the a posteriori distribution of the hidden variable $Z_i$ given the
observation $Y_i$; $q=d$; $\param \mapsto \R(\param) -
\pscal{s}{\phi(\param)}$ possesses an unique minimum; for any $s \in
\Sset \eqdef \rset^d$, $\map(s) = \Omega s$ where
\begin{equation}
  \label{def:omega:example}
\Omega \eqdef \left( \frac{1}{\sigma^2 n}
           \sum_{i=1}^n \frac{X_iX_i^T}{\|X_i\|^2} + 2 \tau \Id_d
           \right)^{-1} \eqsp.
\end{equation}
           
When such a map $\map$ exists, it is well known that EM can be
equivalently defined in the expectation step: the computation of the
$\Param$-valued sequence $\{\param_k, k \geq 0\}$ through
$\param_{k+1}=\map \circ \bars(\param_k)$ is equivalent to the
computation of the $\bars(\Theta)$-valued sequence $\{s_k, k \geq 0
\}$ through $s_{k+1} = \bars \circ\map(s_k)$.  The limiting points of
these sequences are resp. the roots of $\param \mapsto \map \circ
\bars(\param) - \param$ and $s \mapsto \bars \circ \map(s) -s$ (see
e.g. \cite{Delyon:lavielle:moulines:1999}). Hereafter, we will see EM
as an algorithm in the expectation space: EM is an iterative procedure
designed to find the roots of {\em the mean field $\mf$}: $\Sset \to
\rset^q$
\[
\mf(s) \eqdef \bars \circ \map(s) -s = \frac{1}{n} \sum_{i=1}^n
\bars_i \circ \map(s) -s \eqsp.
\]
In the large scale learning setting, $\bars$ has a prohibitive
computational cost since it involves a sum over the full data set of
size $n$: EM can not be applied exactly.  A popular alternative in the
literature is to replace EM iterations with SA iterations, where the
SA algorithm is designed to find the roots of $\mf$
\cite{Delyon:lavielle:moulines:1999}. {\tt 3P-SPIDER} is in the same
vein.

\vspace{-0.2cm}
\subsection{The Perturbed Prox-Preconditioned SPIDER algorithm}
Given a sequence of positive step sizes $\{\pas_k, k \geq 0\}$, SA
defines a sequence $\{\hatS_k, k \geq 0 \}$ such that
\begin{equation}
  \label{eq:SA}
  \hatS_{k+1} = \hatS_k + \pas_{k+1} H_{k+1}
\end{equation}
where $H_{k+1}$ is an approximation of $\mf(\hatS_k)$.  Observing that
$\bars(\param) = \PE\left[\bars_I(\param)\right]$ for some
$[n]^\star$-valued uniform random variable $I$, a natural idea to
mimic the asymptotic behavior of EM is the definition
\[
H_{k+1} \eqdef \frac{1}{\lbatch} \sum_{i \in \batch_{k+1}}
\bars_i(\hatS_k) - \hatS_k
\]
where $\batch_{k+1}$ is a batch of size $\lbatch$ sampled uniformly
from $[n]^\star$ (with or without replacement) and independently of
$\hatS_{k}$.  Such a strategy corresponds to the {\tt Online-EM}
algorithm. The incremental EM-based algorithms with variance reduction
techniques use the property $\mf(\hatS_k) = \PE\left[H_{k+1} + V \vert
  \hatS_k \right]$ for any (conditionally) centered random variable
$V$. This implies that, thanks to an adequate construction of the {\em
  control variate} $V$, the variance of the approximation of
$\mf(\hatS_k)$ can be reduced (see e.g. \cite{glasserman:2004} for an
introduction to variance reduction methods in Monte Carlo
sampling). This is the essence of {\tt sEM-vr}, {\tt FIEM} and {\tt
  SPIDER-EM} which essentially differ in the definition of $V$.

{\tt 3P-SPIDER} is described in Algorithm~\ref{algo:3PSPIDER}.  As in
{\tt SPIDER-EM}, the control variate is refreshed regularly, let us
say at the beginning of each {\em outer} loop $\# t$ (see lines
\ref{eq:SA:reset:0} and \ref{eq:SA:reset:1}). In {\tt SPIDER-EM}, it
is defined as $\bars \circ \map(\hatS_{t,-1}) = n^{-1} \sum_{i=1}^n
\bars_i(\hatS_{t,-1})$. Here, two perturbations are allowed: the
approximation of $\bars_i(\hatS_{t,-1})$ with a quantity denoted by
$\hats^{t,-1}_i$, and an error $\mathcal{E}_t$ which may include for
example the situation when a sub-sample of the $n$ examples is used
when computing the sum instead of the full data set. At each {\em
  inner} loop $\# (k+1)$, the control variate is modified in order to
track the ideal quantity $\bars \circ \map(\hatS_{t,k})$: note indeed
that $\Smem_{t,0} \approx \bars \circ \map(\hatS_{t,-1})$ and, from
line \ref{eq:SA:update:Smem}, $\Smem_{t,k+1} - \Smem_{t,k} \approx
\bars \circ \map(\hatS_{t,k}) - \bars \circ \map(\hatS_{t,k-1})$.

The sequence of interest $\{\hatS_{t,k}, t \in [\kouter]^\star, k \in
[\kin] \}$ is updated first by a SA step (see
Line~\ref{eq:SA:updateclassical}) followed with a proximal step (see
Line~\ref{eq:SA:proximal}). In the SA step, the mean field
$\mf(\hatS_{t,k}) = \bars \circ \map(\hatS_{t,k}) - \hatS_{t,k}$ is
approximated with (see Lines~\ref{eq:SA:update:Smem} and
\ref{eq:SA:updateclassical})
\[
H_{k+1} \eqdef \frac{1}{\lbatch} \sum_{i \in \batch_{t,k+1}}
\hats_i^{t,k} +V_{k+1} - \hatS_{t,k}
\]
where $V_{k+1} \eqdef \Smem_{t,k} - \lbatch^{-1} \sum_{i \in
  \batch_{t,k+1}} \hats_i^{t,k-1}$ is a control variate. Here again,
$\batch_{t,k+1}$ is a batch of size $\lbatch$ sampled from
$[n]^\star$, with or without replacement and independently of the past
of the algorithm.

The proximal step in lines \ref{eq:SA:proximal} and
\ref{eq:SA:proximal-2} is a novelty (with respect to {\tt SPIDER-EM})
introduced to force the path of the algorithm $\{\hatS_{t,k}, t \in
[\kouter]^\star, k \in [\kin] \}$ to remain in the set $\Sset$ and
possibly to inherit other properties from an adequate definition of
$g$ (see section~\ref{sec:logistic} for an example).  The proof of the
convergence in expectation of the algorithm (see
\cite{fort:moulines:SSP21:theory}) relies on the observation that the
algorithm \eqref{eq:SA} is a perturbed {\em preconditioned}-gradient
method: by setting $\lyap(s) \eqdef F \circ \map(s)$, we have under
regularity assumptions on the functions $\phi, \s, \R$ that $\nabla
\lyap(s) = - B(s) \mf(s)$ for any $s \in \Sset$, where (see
e.g. \cite[Proposition 1]{fort:moulines:gach:2020})
\[
B(s) \eqdef \left( \nabla \map(s)\right)^T \,\nabla^2_\param \left(
\R(\param) - \pscal{s}{\phi(\param)} \right) \vert_{\param = \map(s)} \ \left( \nabla \map(s)
\right) \eqsp,
\]
is a positive-definite matrix.  Therefore, given a lower
semi-continuous proper convex function $g: \Sset \to \rset\cup \{+
\infty\}$, we use a {\em weighted} proximal operator defined by
\[
\Prox_{B,\pas g}(s') \eqdef \mathrm{argmin}_{s \in \Sset} \left(\pas
g(s) +\frac{1}{2} (s-s')^T B (s-s')\right)
\]
for any $\pas>0$ and any $q \times q$ positive-definite matrix $B$.

{\tt 3P-SPIDER} extends {\tt SPIDER-EM} in the following directions.
First, in the definition of the control variates $\Smem_{t,k}$, it
allows to substitute the intractable $\bars_i \circ \map(\hatS_{t,k})$
with an approximation $\hats^{t,k}_i$. Second, it adds a proximal
step in order to force the sequence $\{\hatS_{t,k}, t \in
[\kouter]^\star, k \in [\kin]\}$ to have some properties (see
\cite{li:li:2018,wang:etal:2019} for a similar idea applied to the
     {\tt SPIDER} algorithm, with $B(s) =\Id_d$). Finally, it allows a
     perturbation $\mathcal{E}_t$ when initializing the control
     variate $\Smem_{t,0}$.

\begin{algorithm}[htbp]
    \caption{The Perturbed Prox-Preconditioned SPIDER (3P-SPIDER)
      algorithm.
      \label{algo:3PSPIDER}}
   \KwData{ $\kouter, \kin \in \nset^\star$; $\hatS_\init \in \Sset$;
     $\pas_{t,0} \geq 0$, $\pas_{t,k} >0$ for $t \in [\kouter]^\star$,
     $k \in [\kin]^\star$, a lower semi-continuous proper convex
     function $g$} \KwResult{The 3P-SPIDER sequence $\{\hatS_{t,k}, t
     \in [\kin]^\star, k \in [\kin]\}$} $\hatS_{1,0} = \hatS_{1,-1} =
   \hatS_\init$ \; $\Smem_{1,0} = n^{-1} \sum_{i=1}^n \hats^{1,-1}_i+
   \mathcal{E}_1$ \label{eq:SA:reset:0} \; \For{$t=1, \cdots,
     \kouter$ \label{eq:SA:epoch}}{ \For{$k=0, \ldots,\kin-1$}{Sample
       a mini batch $\batch_{t,k+1}$ of size $\lbatch$ in
       $[n]^\star$ \label{eq:SA:update:batch} \; $\Smem_{t,k+1} =
       \Smem_{t,k} + \lbatch^{-1} \sum_{i \in \batch_{t,k+1}} \left(
       \hats_i^{t,k} - \hats_i^{t,k-1} \right)
       $ \label{eq:SA:update:Smem} \; $\hatS_{t,k+1/2} = \hatS_{t,k} +
       \pas_{t,k+1} \left( \Smem_{t,k+1} - \hatS_{t,k}
       \right)$ \label{eq:SA:updateclassical} \; $\hatS_{t,k+1} =
       \Prox_{B(\hatS_{t,k}), \pas_{t,k+1} g}\left( \hatS_{t,k+1/2}
       \right)$ \label{eq:SA:proximal} \;} $\hatS_{t+1,-1} =
     \hatS_{t,\kin}$ \; $\Smem_{t+1,0}= n^{-1} \sum_{i=1}^n
     \hats_i^{t+1,-1} + \mathcal{E}_{t+1}$ \label{eq:SA:reset:1} \;
     $\hatS_{t+1,-1/2} = \hatS_{t+1,-1} + \pas_{t+1,0} \left(
     \Smem_{t+1,0} - \hatS_{t+1,-1} \right)$ \; $\hatS_{t+1,0} =
     \Prox_{B(\hatS_{t+1,-1}), \pas_{t+1,0}
       g}(\hatS_{t+1,-1/2})$ \label{eq:SA:proximal-2} }
\end{algorithm}
In \cite{fort:moulines:2021} (see also \cite{fort:moulines:SSP21:theory}), the convergence in expectation of the
sequence $\{\hatS_{t,k}, t \in [\kouter]^\star, k \in [\kin] \}$
towards the set
\begin{align*}
  \mathcal{L} &\eqdef \{s: \Prox_{B(s), \pas g}(s + \pas \mf(s)) = s
  \} \qquad \forall \pas >0 \eqsp,\\ &= \{s: 0 \in \partial g(s) -
  B(s) \mf(s) \} = \{s: 0 \in \partial g(s) + \nabla \lyap(s) \}
  \end{align*}
is proved. In the case $g$ is the indicator function of a closed
convex set $\mathcal{K}$ and $B(s)$ is invertible for any $s \in
\mathcal{K} \cap \Sset$, the limiting points are the roots of $\nabla
\lyap$ which are in $\mathcal{K}$, that is the roots of $\mf(s)$ in
$\mathcal{K}$: {\tt 3P-SPIDER} has the same asymptotic behavior as EM.

\subsection{Case of a Monte Carlo approximation}
\label{sec:MCapprox}
The intractable quantity $\bars_i \circ \map(s)$ is defined by
\[
\bars_i \circ \map(s) \eqdef \int_\Zset \s_i(z) p_i(z; \map(s))
\rmd \mu(z) \eqsp,
\]
where $p_i(z; \map(s)) \, \rmd \mu(z)$ is the distribution defined by
\eqref{eq:aposteriori}. When this integral is not explicit, a natural
idea is to approximate it by a Monte Carlo (MC) sum. For example,
\[
\hats^{t,k}_i \eqdef \frac{1}{m_{t,k+1}} \sum_{r=1}^{m_{t,k+1}}
\s_i\left( Z_{r}^{i,t,k} \right) \eqsp,
\]
where for $t \in [\kouter]^\star, k \in [\kin]$ and $i \in [n]^\star$,
$\{Z_r^{i,t,k}, r \geq 1 \}$ is a Markov chain designed to be ergodic
with unique invariant distribution $p_i(z; \map(\hatS_{t,k})) \rmd
\mu(z)$. Such a chain can be obtained by running a Markov chain Monte
Carlo sampler (see e.g. ~\cite{cappe:robert:2000,MCMCbook:2011}); note
that the independent and identically distributed (i.i.d) setting is a
special case of the Markovian setting. When the random variables
$\{Z_r^{i,t,k}, r \geq 1 \}$ are i.i.d., we have
$\PE\left[\hats^{t,k}_i \vert \hatS_{t,k} \right] = \bars_i \circ
\map(\hatS_{t,k})$; when the random variables $\{Z_r^{i,t,k}, r \geq 1
\}$ are a Markov chain, the approximation is biased:
$\PE\left[\hats^{t,k}_i \vert \hatS_{t,k} \right] \neq \bars_i \circ
\map(\hatS_{t,k})$. In this biased case, the algorithm still converges
to $\mathcal{L}$ but its theoretical analysis is more technical (see
\cite{fort:moulines:2021}).

\section{Application: Inference in the Logistic Regression Model}
\label{sec:logistic}
Let us consider the logistic regression model described in
Section~\ref{sec:optimization}. Since $\PP(Y_i = y_i) \leq 1$, it can
be proved that the minima of $F$ are in the set $\{\param \in \rset^d:
\tau \|\param\|^2 \leq \ln 4 \}$. Therefore, in the expectation space,
the minima of $\lyap = F \circ \map$ are in the set $\{s \in \rset^d:
\tau s^T \Omega^2 s \leq \ln 4 \}$ which is included in $\mathcal{K}
\eqdef \{s \in \rset^d: \tau s^T \Omega s \leq \ln 4 /\lambda_{\min}
\}$ where $\lambda_{\min}$ is the positive minimal eigenvalue of
$\Omega$ (see \eqref{def:omega:example}).  {\tt 3P-SPIDER} is applied
with $g \eqdef \indic_{\mathcal{K}}$, the characteristic function of
the compact convex set $\mathcal{K}$. With this definition of
$\mathcal{K}$ and since $B(s) = \Omega$ for any $s \in \Sset
\eqdef\rset^d$, the computation of the operator $\Prox_{B(s), \pas g}$
is explicit. $\bars_i(\Omega s)$ is equal to
\begin{multline*}
\frac{X_i}{\sigma^2 \|X_i\|} \frac{1}{Z(s)} \, \int_\rset z \,
\frac{\exp(z \pscal{X_i}{\Omega s}/(\sigma^2 \|X_i\|))}{1+ \exp(- Y_i
  \|X_i\| z)} \, \exp(\frac{-z^2}{2 \sigma^2}) \rmd z
\end{multline*}
where the normalizing constant $Z(s)$ is given by
\[
Z(s) \eqdef \int_\rset \frac{\exp(z \pscal{X_i}{\Omega s}/(\sigma^2
  \|X_i\|))}{1+ \exp(- Y_i \|X_i\| z)} \, \exp(-z^2/(2 \sigma^2)) \rmd
z \eqsp.
\]
These integrals are not explicit; we consider the approximation
$\hats_i^{t,k}$ of $\bars_i(\Omega \hatS_{t,k})$ given by a MC sum as
described in Section~\ref{sec:MCapprox}; the samples $Z_r^{i,t,k}$ are
obtained by the Gibbs sampler given in
\cite{polson:scott:windle:2013}.

The numerical illustrations use the MNIST data set: the class "$1$"
contains the $12 \, 873$ images in the training set labeled $1$ and
$3$ and the class "$-1$" contains the $12 \, 116$ images in the
training set labeled $7$ and $8$; hence $n=24 \, 989$. The $787$
pixels are compressed in $50$ features through PCA (see \cite[section
  5]{fort:moulines:gach:2020} for the details). An intercept is
included in the covariates so $d=51$.  {\tt 3P-SPIDER} is run with
$\sigma^2=0.1$, $\tau =1$, $\kouter = 20$, $\kin = \lceil \sqrt{n}/10
\rceil =16$ and $\lbatch = \lceil 10 \sqrt{n} \rceil=1581$. Note that
$\kin \times \lbatch =n$ so that each outer loop requires $n$
examples; it corresponds to an {\em epoch}.

We study the quantity
\[
\mathcal{D}_{t,k} \eqdef \PE\left[ \frac{\|\hatS_{t,k+1} -
    \hatS_{t,k}\|^2}{\pas_{t,k+1}^2} \right], \qquad t \in
        [\kouter]^\star, k \in [\kin-1],
\]
which quantifies how far {\tt 3P-SPIDER} is from its limiting set (see
the definition of $\mathcal{L}$); this expectation is estimated by a
MC sum over $25$ independent runs. All the runs start from the same
value $\hatS_\init$. For the computation of the quantity $\Smem_{t,0}$
at each outer loop $\# t$, all the examples are used and the
expectations $\bars_i \circ \map(\hatS_{t,-1})$ are approximated by a
MC sum with $m' = 10 \lceil\sqrt{n} \rceil = 1590$ points. Hence,
$\mathcal{E}_t=0$ except when specified (see the second analysis
below); the computation of $\Smem_{t,0}$ requires $n$ examples: it
corresponds to an epoch.

\begin{figure}[h] \centering
\begin{minipage}[b]{0.48\linewidth}
 \includegraphics[width=4.6cm]{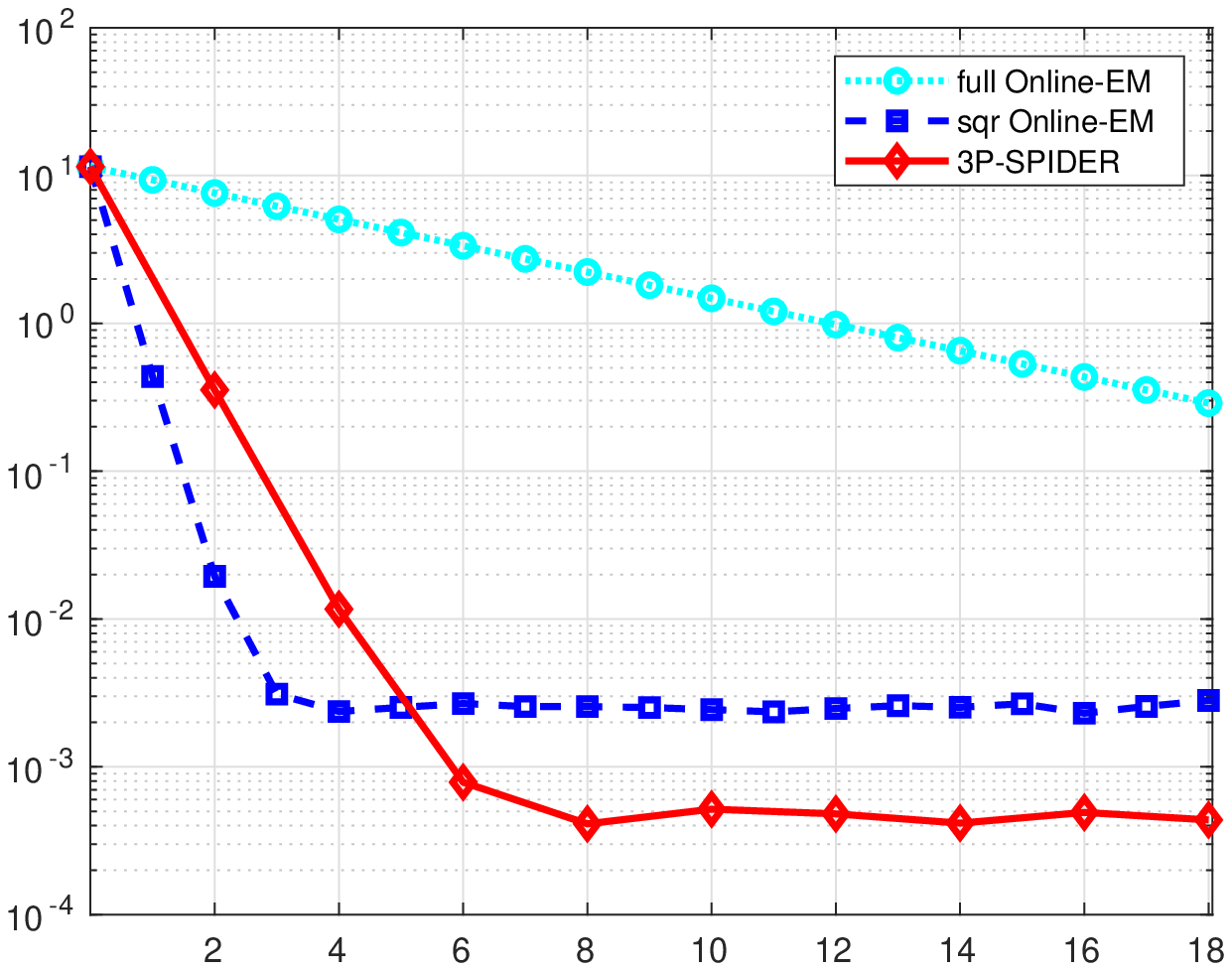} \\
 \includegraphics[width=4.6cm]{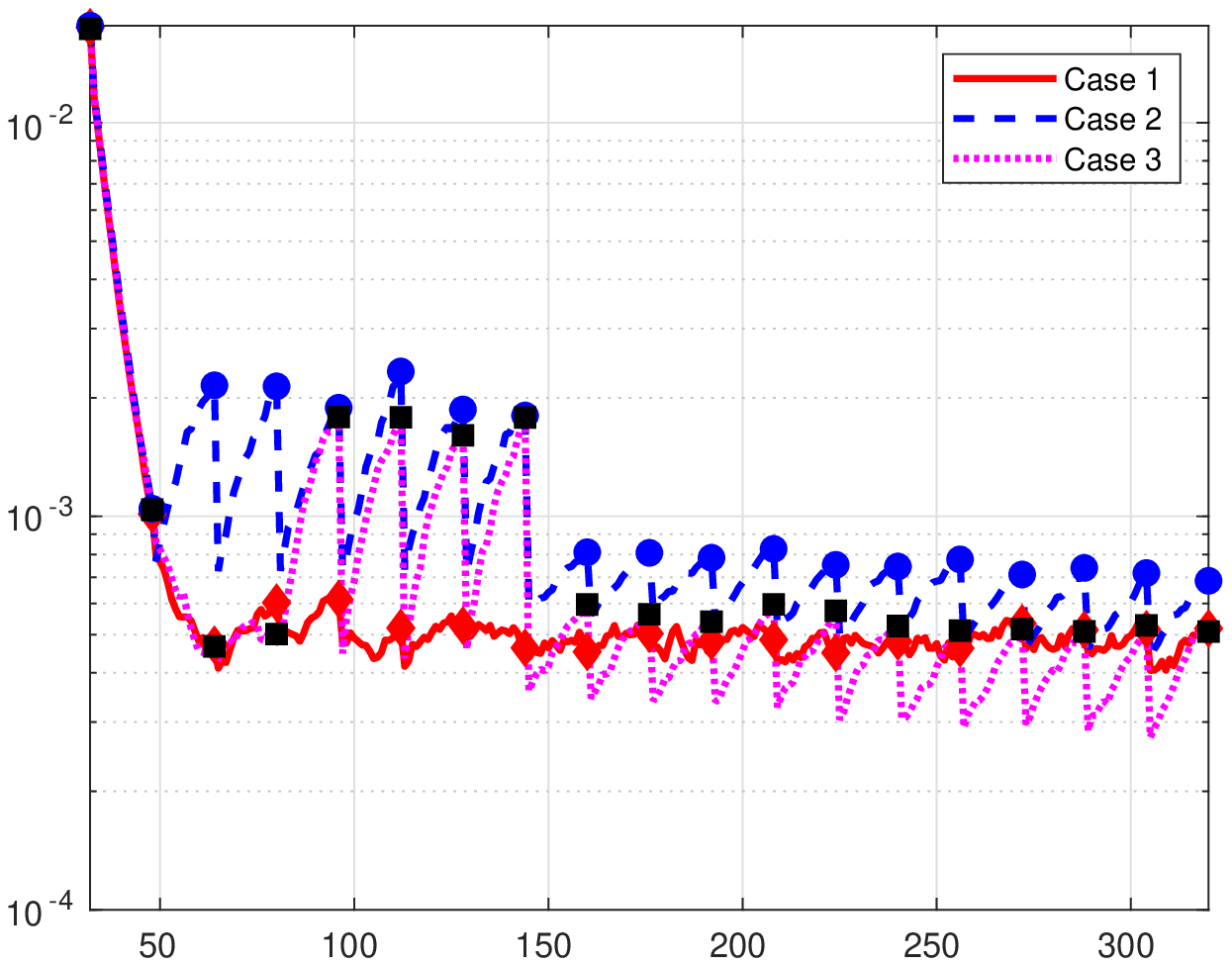}
 \end{minipage}
\hfill
 \begin{minipage}[b]{0.48\linewidth}
  \includegraphics[width=4.6cm]{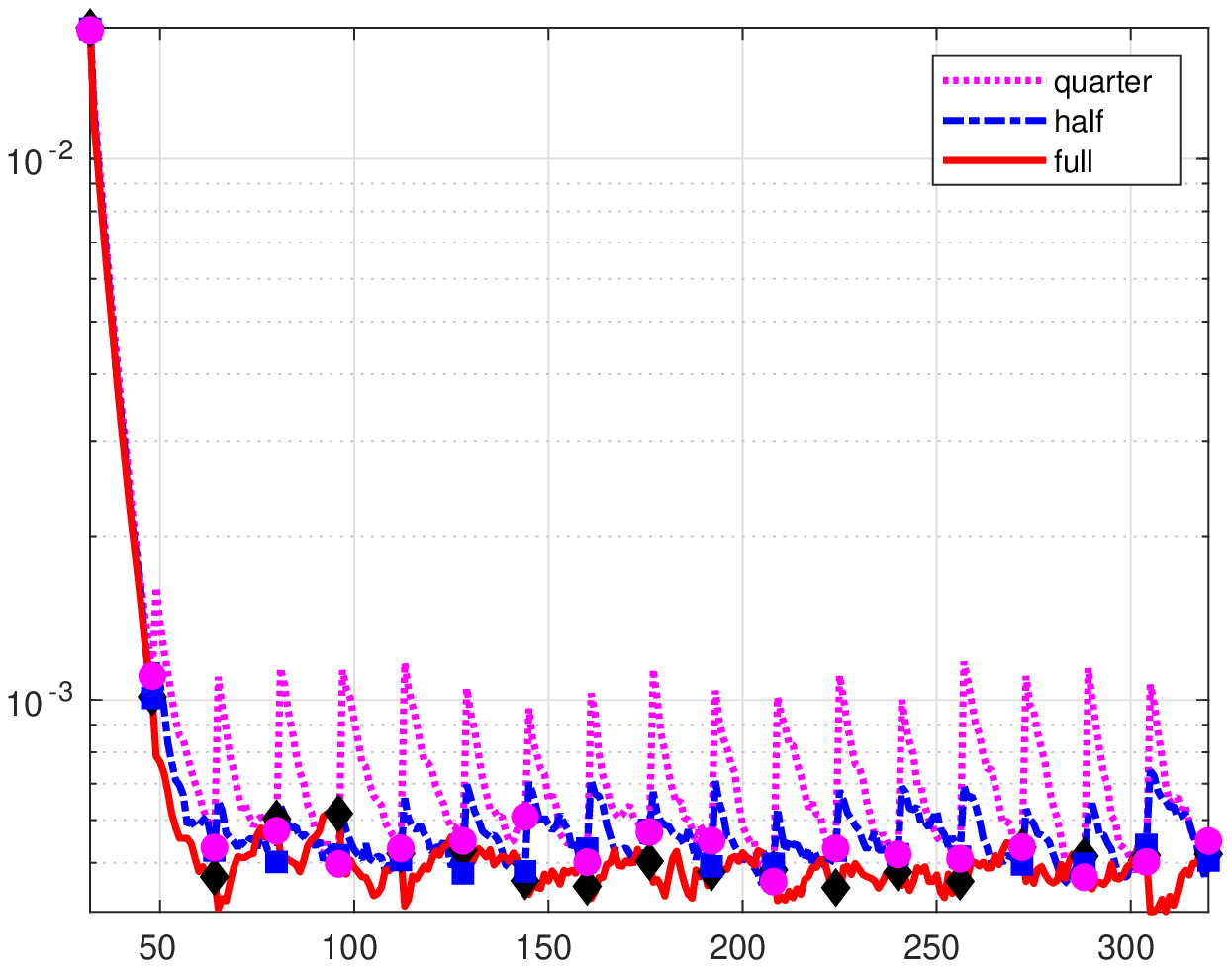} \\
  \includegraphics[width=4.6cm]{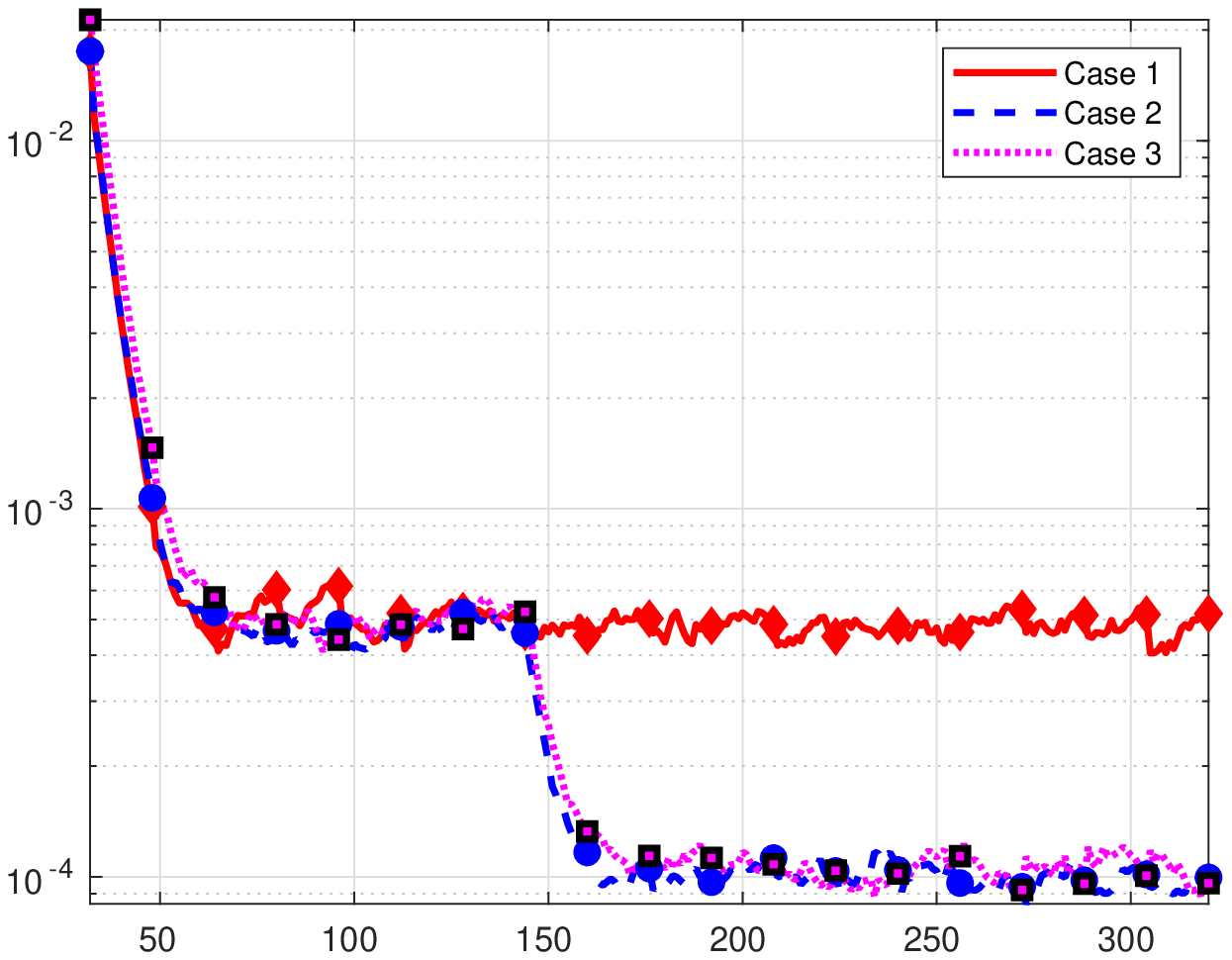}
\end{minipage}
\caption{[(a) top left] Comparison of algorithms; [(b) top right] Role
  of the size of the batch when computing $\Smem_{t,0}$; [(c) bottom
    left] Role of the step sizes $\pas_{t,k}$ and the number of Monte
  Carlo points when computing $\hats^i_{t,k}$; [(d) bottom right] Role
  of $\pas_{t,0}$}
\label{fig:plot}
\end{figure}

First, {\tt 3P-SPIDER} is run with $m_{t,k} = 2 \lceil \sqrt{n}
\rceil$ and $\pas_{t,k} = \pas_{t,0} = 0.1$; {\tt Online-EM} is run
with a step size equal to $\pas_t = 0.1$, a batch size $\lbatch =
\lceil 10 \sqrt{n} \rceil$ (case "sqr") and $\lbatch =n$ (case
"full"), and a MC approximation for $\bars_i$ computed with $2 \lceil
\sqrt{n} \rceil$ points.  Figure~\ref{fig:plot}(a) displays
$\mathcal{D}_{t,0}$ for {\tt 3P-SPIDER} and $\| \hatS_{t+1} -
\hatS_{t}\|^2 /\pas_{t}^2$ for {\tt Online-EM}. The x-axis scales as
the number of {\em epoch}, that is the use of $n$ examples. The plot
shows that, even when the expectations $\bars_i$ have to be replaced
with approximations, {\tt 3P-SPIDER} is far more efficient than {\tt
  Online-EM} (in which the exact expectations are also replaced with
MC approximations).

Second, we analyze the role of $\mathcal{E}_t$ when initializing the
control variate $\Smem_{t,0}$. We run {\tt 3P-SPIDER} with $\pas_{t,k}
= \pas_{t,0}= 0.1$ and a number of MC points $m_{t,k} = 2 \lceil
\sqrt{n} \rceil$; the quantity $\mathcal{D}_{t,k}$ is displayed on
Figure~\ref{fig:plot}(b) vs the cumulated number of inner loops; the
squares, circles and diamonds indicate $\mathcal{D}_{t,\kin}$ for
every outer loop. The case "full" corresponds to $\mathcal{E}_t=0$,
the case "half" (resp. "quarter") corresponds to $\Smem_{t,0}$
computed with a batch of size $\lceil n/2 \rceil$ examples
(resp. $\lceil n/4 \rceil$). The control variate is too poor in the
case "half" and "quarter" and, after the transient phase when the
possibly bad initialization is forgotten, it weakens the benefit of
its use: we definitely advice $\mathcal{E}_t =0$.

Third, we analyze how the variability of the MC approximation and the
choice of the step sizes affect the rate of convergence of {\tt
  3P-SPIDER}. In "Case 1", the values are the same as in
Figure~\ref{fig:plot}(a). In "Case 2", $\pas_{t,k} = \pas_{t,0} = 0.1
$ during the first three outer loops and then $\pas_{t,k} = \pas_{t,0}
=10^{-3}$; $m_{t,k}$ is as in "Case 1" until the outer loop $\# 10$ and
then $m_{t,k}$ is multiplied by $5$. In "Case 3", the step sizes and
the number of MC points are as in "Case 2", except that the step size
decreases later, at outer loop $\# 6$.  On Figure~\ref{fig:plot}(c),
we display $\mathcal{D}_{t,k}$ vs the cumulated number of inner loops,
starting from the number $\# 32$ (that is, at the end of the second
outer loop); the diamonds, circles and squares indicate
$\mathcal{D}_{t,\kin}$. First, {\tt 3P-SPIDER} is improved when the
number of MC points increases; when the fluctuations of the algorithm
is of the same order as the fluctuations of the MC errors, {\tt
  3P-SPIDER} can not go forward anymore in order to reach a more
precise estimation of the parameter (compare "Case 1" and "Case
3"). Small step sizes penalize the algorithm (compare "Case 1" and
"Case 2").

Finally, we discuss the strategies $\pas_{t,0} = 0$ and $\pas_{t,0}
\neq 0$. {\tt 3P-SPIDER} run as in Figure~\ref{fig:plot}(a)
corresponds to "Case 1". In "Case 2" and "Case 3", the number of MC
points is multiplied by $5$ from the outer loop $\# 11$, and
$\pas_{t,k} = 0.1$ for any $k>0$.  In "Case 1" and "Case 2",
$\pas_{t,0} = 0.1$ and in "Case 3", $\pas_{t,0} =0$. On
Figure~\ref{fig:plot}(d), we display $\mathcal{D}_{t,k}$ vs the
cumulated number of inner loops, starting from the loop $\# 32$; the
diamonds, circles and squares indicate $\mathcal{D}_{t,\kin}$. Here
again, we observe the benefit of reducing the MC variability by
increasing the number of MC points (compare "Case 1" to the other
cases); "Case 2" and "Case 3" are almost similar, maybe with a
slightly better behavior for "Case 2".
\vspace{-0.2cm}

\end{document}